\RequirePackage{fix-cm} 
\documentclass[natbib,smallcondensed]{svjour3}    
\smartqed% flush right qed marks, e.g. at end of proof
\usepackage{graphicx}
\usepackage{booktabs}
\usepackage{multirow}
\usepackage{hhline}
\usepackage{enumitem}
\usepackage{amssymb,amsmath,amsfonts,enumerate}
\usepackage{subcaption}
\usepackage[ruled,norelsize,linesnumbered]{algorithm2e}
\usepackage{bbm}
\usepackage[misc,geometry]{ifsym} 
\begin{document}

\title{Leveraging Cross Feedback of User and Item Embeddings with Attention for Variational Autoencoder based Collaborative Filtering
}

\author{Yuan Jin         \and
        He Zhao \and Ming Liu \and Ye Zhu \and Lan Du \and Longxiang Gao \and He Zhang \and Yunfeng Li %etc.
}

\institute{ Yuan Jin   \at  CNPIEC KEXIN LTD \\
              \email{jinyuan@kxsz.net}      
           \and
         He Zhao   \at  Faculty of Information Technology, Monash University, Clayton, Victoria, 3168, Australia \\
              \email{ethan.zhao@monash.edu}      
           \and
         Ming Liu   \at  School of Information Technology, Deakin University, Burwood, Victoria, Australia 3125.  \\
              \email{m.liu@deakin.edu.au}      
           \and
 Ye Zhu \at
           School of Information Technology, Deakin University, Burwood, Victoria, Australia 3125.  \\
              \email{ye.zhu@ieee.org}     \and
        Lan Du   \at  Faculty of Information Technology, Monash University, Clayton, Victoria, 3168, Australia \\
              \email{Lan.Du@monash.edu}      
           \and
        Longxiang Gao   \at   School of Information Technology, Deakin University, Burwood, Victoria, Australia 3125.  \\
              \email{longxiang.gao@deakin.edu.au}      
           \and
        He Zhang   \at  CNPIEC KEXIN LTD  \\
              \email{zhanghe@kxsz.net}      
           \and
        Yunfeng Li   \at  CNPIEC KEXIN LTD  \\
              \email{liyunfeng@kxsz.net}   %  \\%      
}

\date{Received: date / Accepted: date}

\maketitle

\begin{abstract}
Matrix factorization (MF) has been widely applied to collaborative filtering in recommendation systems. Its Bayesian variants can derive posterior distributions of user and item embeddings, and are more robust to sparse ratings. However, the Bayesian methods are restricted by their update rules for the posterior parameters due to the conjugacy of the priors and the likelihood. Variational autoencoders (VAE) can address this issue by capturing complex mappings between the posterior parameters and the data. However, current research on VAEs for collaborative filtering only considers the mappings based on the explicit data information while the implicit embedding information is overlooked.

In this paper, we first derive evidence lower bounds (ELBO) for Bayesian MF models from two viewpoints: \textit{user-oriented} and \textit{item-oriented}. Based on the ELBOs, we propose a VAE-based Bayesian MF framework. It leverages not only the data but also the embedding information to approximate the user-item joint distribution. As suggested by the ELBOs, the approximation is iterative with cross feedback of user and item embeddings into each other's encoders. More specifically, user embeddings sampled at the previous iteration are fed to the item-side encoders to estimate the posterior parameters for the item embeddings at the current iteration, and vice versa. The estimation also attends to the cross-fed embeddings to further exploit useful information. The decoder then reconstructs the data via the matrix factorization over the currently re-sampled user and item embeddings. We show the effectiveness of our framework in lowering rating prediction errors, improving relevant item ranking and handling rating sparsity across five real-world datasets. We also perform ablation studies and convergence analysis~to illustrate the importance of the cross-feedback and attention components of our framework in the performance and the model convergence. 
\end{abstract}

\section{Introduction}
Collaborative filtering (CF) approaches~\citep{koren2015advances} have been the backbone of modern recommendation systems. Their goal
is to suggest new items to a particular user that are of interest to him/her based on ratings from other like-minded users or various types of side information (e.g. rating timestamps and locations, user demographics, item content). Among them, matrix factorization (MF) has been the most widely applied approach~\citep{koren2009matrix}. It aims to reconstruct observed ratings from two sets of multi-dimensional embeddings: one for users and the other for items. The user and item embeddings are usually estimated by minimizing the regularized squared (reconstruction) errors via stochastic gradient descent. During the estimation, the embeddings of like-minded users tend to be mapped into close proximity. As a result, a user tend to be recommended with new items that are liked by those with similar embeddings.  %When a new item is recommended to a user, the MF models

From a probabilistic standpoint, the regularization in the above minimization problem is equivalent to imposing Normal priors over the embeddings. The squared errors between the observed and reconstructed ratings are equivalent to Normal likelihood. Thus, the entire problem can also be viewed as maximizing a posteriori over the embeddings~\citep{mnih2008probabilistic}. A further treatment is to make fully Bayesian modelling and inference of the embeddings in matrix factorization. This is referred to as the Bayesian matrix factorization (BMF) for collaborative filtering.

In this case, the BMF models impose appropriate prior distributions over the user and item embeddings~\citep{salakhutdinov2008bayesian,lakshminarayanan2011robust,harvey2011bayesian}. These embeddings then construct a likelihood term over the rating data. Training BMF models amounts to infer the joint posterior distribution of the embeddings analytically or by approximation. In most cases, distributions involving multiple variables cannot be derived analytically, and thus approximation inference techniques are used instead. 

BMF models adopt two types of approximation inference frameworks: Markov Chain Monte Carlo (MCMC) methods and Variational Inference. Both of them rely on the \textit{conjugacy} of the priors and the likelihood over the embeddings. The BMF models have mostly exploited Normal-Normal and Poisson-Gamma conjugacy for their likelihood and priors. Despite its algebraic convenience, the conjugate updates on the posterior parameters of each embedding are independent and thus fail to utilize the information from each other. This limits the performance of the BMF models. In this case, we want the data information for different embeddings to be shared to refine the inference of their posterior distributions.

To enable the sharing of the data information, it is common to model the posterior parameters as regression over the data. The regression weights are learned to map similar data patterns into values in close proximity in the latent space. Naturally, the mapping is non-linear. This motivates us to use artificial neural networks (NN), which can fit arbitrarily complex mapping functions, to predict the posterior parameters for the embeddings. In particular, \textit{variational auto-encoders} (VAEs) provide the foundation to achieve all of the above. It connects variational inference of posterior distributions with neural networks. 

Motivated by our derivation of two ELBOs for the variational inference of BMF models in collaborative filtering, we propose a novel BMF framework based on variational autoencoders. It conducts \textit{mean-field} variational inference for user and item embeddings by predicting their posterior parameters using multi-layer perceptron (MLP) encoders. These encoders take in not only the data but also the information from the embeddings. Such an input arrangement is evidenced by the inter-dependence of variables in the derived ELBOs; that is a user's embeddings dependent on both his/her ratings and the embeddings of the rated items, and vice versa.  

Our framework leverages both the data information and the embedding information in an iterative manner. In each iteration, user and item embeddings are sampled from their respective encoders. The user-side encoders take in the rating inputs and latent inputs popluated by the item embedding samples which are fed back from the last iteration, and vice versa. In addition, we enable the encoders to attend to certain parts of the latent inputs to further exploit useful information from the embeddings. Finally, the decoder reconstructs the ratings using the matrix factorization over the currently (re-)sampled user and item embeddings.

The contributions made by our paper are listed as follows:
\begin{itemize}
\item We derive \textit{user-oriented} and \textit{item-oriented} ELBOs for the varitional inference of BMF models in collaborative filtering.
\item We propose a VAE-based BMF framework that unifies the modelling of both derived ELBOs. It achieves overall lower reconstruction errors and higher relevant item ranking measures compared to six state-of-the-art NN-based CF methods across five real-world rating datasets.
\item When the datasets are subsampled to emulate the sparsity that might occur across both users and items, we found that our framework which leverages the implicit embedding information is overall more robust than the baseline models.
\item The cross-feedback and attention components clearly improve the performance and the convergence of our framework, as revealed by an ablation study and a model convergence analysis.
\end{itemize}

\section{Related Work}
Matrix factorization (MF)~\citep{koren2015advances} is the most widely applied approach in collaborative filtering (CF). It assumes that a matrix can be well approximated by the multiplication of two low-rank matrices. In the context of CF, the target matrix stores the user-item ratings, while the two low-rank matrices comprise the latent representations, called the embeddings, of users and items. Based on the MF modelling, regularized squared errors between the observed entries (stored in the rating matrix) and the approximated ones are minimized with respect to the user and item embeddings. This minimization is usually conducted via stochastic gradient descent (SGD)~\citep{koren2009matrix,koren2015advances}. Among various MF models, the probabilistic matrix factorization (PMF)~\citep{mnih2008probabilistic} proposed to assign Normal priors to the user and item embeddings. These embeddings further form a Normal likelihood over the ratings. Each rating is modelled to follow a Normal distribution centered on a user-item embedding dot product. The above minimization problem is now formulated as maximum a posteriori (MAP) estimation of the embeddings via SGD. However, both of these cases yield point estimates for the embeddings. 

\subsection{Bayesian Matrix Factorization}
Extending from the PMF model are the Bayesian matrix factorization (BMF) models \citep{salakhutdinov2008bayesian,Gopalan_2015,Gopalan2015SRH}. They impose various priors over the user and item embeddings. These priors are usually conjugate to the likelihood over the rating data. In~\citep{salakhutdinov2008bayesian}, a fully Bayesian treatment was applied in which Normal-Wishart priors were imposed over the embeddings. Due to the conjugacy of such priors and a Normal likelihood,
the conditional posterior distributions of the embeddings can be sampled in a closed form via the Gibbs sampling. 

\citet{lakshminarayanan2011robust} replaced the Normal priors with some heavy-tailed distributions such as the Student-t distributions to increase the robustness of BMF to outlier or atypical ratings. \citet{Gopalan_2015,Gopalan2015SRH} proposed to impose Gamma distributions as the priors over the embeddings. In this case, the likelihood is assumed to follow a Poisson distribution which exploits the Poisson-Gamma conjugacy for the convenience of following inference. All these work leverages the variational inference~\citep{blei2017variational} to estimate the posterior distribution parameters for the embeddings. %The conjugate priors employed enable the inference but the update rules they exert on the posterior parameters are very often restricted in capturing useful data information.

\subsection{Variational Inference}
According to~\citet{blei2017variational}, the goal of variational inference is to approximate some intractable conditional distribution of latent variables $\boldsymbol{Z}$ given observations $\boldsymbol{X}$: $p(\boldsymbol{Z}|\boldsymbol{X})$. The approximation amounts to solving an optimization problem over a family of tractable distribution $q(\boldsymbol{Z})$. The objective function in this case is the Kullback-Leibler (KL) divergence between the true intractable distribution $p(\boldsymbol{Z}|\boldsymbol{X})$ and the tractable variational one $q(\boldsymbol{Z})$: $\text{KL}\big[q(\boldsymbol{Z})||p(\boldsymbol{Z}|\boldsymbol{X})\big]$. It is minimized with respect to the distribution parameters of $q(\boldsymbol{Z})$, which turns out to be equivalent to maximizing the following evidence lower bound (ELBO) Q:
\begin{equation}
\begin{split}
\text{Q}&=\mathbbm{E}_{q}\big[\log{p(\boldsymbol{X},\boldsymbol{Z})}\big] -\mathbbm{E}_{q}\big[\log{q(\boldsymbol{Z})}\big]\hspace{10pt}=\mathbbm{E}_{q}\big[\log{p(\boldsymbol{X}|\boldsymbol{Z})}\big] - \text{KL}\big[q(\boldsymbol{Z})||p(\boldsymbol{Z})\big]\label{eqn:elbo}
\end{split}
\end{equation}
In Eq.~\ref{eqn:elbo}, $\mathbbm{E}_{q}\big[\log{p(\boldsymbol{X},\boldsymbol{Z})}\big]$ and $\mathbbm{E}_{q}\big[\log{p(\boldsymbol{x}|\boldsymbol{Z})}\big]$ are respectively the expected joint distribution and data log-likelihood with respect to (the parameters of) $q(\boldsymbol{Z})$ and $\text{KL}\big[q(\boldsymbol{Z})||p(\boldsymbol{Z})\big]$ is the KL divergence between $q(\boldsymbol{Z})$ and the prior distribution $p(\boldsymbol{Z})$.

The variational distribution $q(\boldsymbol{Z})$ needs to be easy to optimize over. One such family of distributions is the mean-field distributions which are the set of fully factored independent distributions from $q(\boldsymbol{Z})$: $q(\boldsymbol{Z})=\prod_{n=1}^{N} q(\boldsymbol{z}_n)$ where $N$ is the size of the data. In this case, the ELBO function Q in Eq.~\ref{eqn:elbo} can be simplified and expanded as follows:
\begin{equation}
    \text{Q}=\sum_{n=1}^{N}\mathbbm{E}_{q}\big[\log{p(\boldsymbol{x}_n|\boldsymbol{z}_n)}\big] - \sum_{n=1}^{N}\text{KL}\big[q(\boldsymbol{z}_n)||p(\boldsymbol{z}_n)\big]\label{eqn:elbo1}
\end{equation}
In collaborative filtering, it is conventional to set each $q(\boldsymbol{z}_n)$ and $p(z_n)$ to be Normal distributions: $q(\boldsymbol{z}_n)=\mathcal{N}(\boldsymbol{\mu}_n,\text{diag}(\boldsymbol{\sigma}^{2}_n))$ and $p(\boldsymbol{z}_1)=p(\boldsymbol{z}_2)=...=p(\boldsymbol{z}_n)=\mathcal{N}(\boldsymbol{\mu},\text{diag}(\boldsymbol{\sigma}^{2}))$. Maximizing the ELBO with respect to each parameter $\{\boldsymbol{\mu}_n,\text{diag}(\boldsymbol{\sigma}^{2}_n)\}_{n=1,...,N}$ via the coordinate ascent~\citep{blei2017variational} yields iterative posterior updates on the parameters. This means that updates on $\boldsymbol{\mu}_n$ and $\text{diag}(\boldsymbol{\sigma}^{2}_n)$ take in the data information either directly from $\boldsymbol{x}_n$ or indirectly from each other. 

One major problem of the update rules based on the variational inference is their lack of flexibility to capture the full complexity of the true posterior distribution $p(\boldsymbol{Z}|\boldsymbol{X})$~\citep{zhang2018advances}. This is largely due to the conjugate pairs (e.g. Normal-Normal, Poisson-Gamma, Multinomial-Dirichlet) imposed by the variational models on the likelihood and priors. Furthermore, the update for the variable $\boldsymbol{z}_n$ fails to leverage the update information for the other variables $\{\boldsymbol{z}_{n'}\}_{1 \leq n' \leq N, n' \neq n}$. 
\subsection{Variational Autoencoder Based Bayesian Matrix Factorization}
%Autoencoders~\citep{pmlr-v27-baldi12a} (AE) are a family of unsupervised generative neural networks that reconstruct their input data and meanwhile generate embeddings of the input data. Among them, 
Variational autoencoders (VAE)~\citep{kingma2013auto} can solve both of the above problems. They are extended from autoencoders (AE)~\citep{pmlr-v27-baldi12a} which are unsupervised generative neural networks. They aim to reconstruct the input data and meanwhile generate multi-dimensional embeddings of the data. VAEs are known to be able to approximate intractable posterior distributions of the embeddings. This is achieved via computing the variational posterior parameters of the embeddings using particular neural networks (e.g. MLPs). These neural networks are referred to as the encoders of the VAEs. More specifically, the variational posterior distribution $q(\boldsymbol{z}_n)$ of the multi-dimensional embedding $\boldsymbol{z}_n$ are:
\begin{equation}
   q(\boldsymbol{z}_n)=\mathcal{N}\big(\boldsymbol{\mu}_{n},\text{diag}(\boldsymbol{\sigma}^2_{n})\big)=\mathcal{N}\big(\boldsymbol{F}(\boldsymbol{x}_n), \boldsymbol{F}'(\boldsymbol{x}_n) \big)~\label{eqn:posterior_sample}
\end{equation}
In Eq.~\ref{eqn:posterior_sample}, $\boldsymbol{F}$ and $\boldsymbol{F}'$ are the non-linear functions modelled by the encoder neural networks. In this case, the neural networks take in each data point $\boldsymbol{x}_n$ and output the corresponding mean vector $\boldsymbol{\mu}_{n}$ and the variance vector $\text{diag}(\boldsymbol{\sigma}^2_{n})$.

VAEs can capture arbitrarily complex mapping relationships between the posterior parameters and the data, which failed to be captured by the conjugate updates of the BMF models. VAEs also allow the sharing of the posterior update information through the learned encoder weights while the update information are independent in the BMF models. 

In recent years, research has started to emerge which integrates VAEs into the BMF in collaborative filtering. \citet{li2017collaborative} proposed to leverage VAEs for estimating the posterior means and variances of item embeddings from the item content. Meanwhile, the generated item embeddings are involved in the traditional BMF modelling. \citet{liangvae2018} only estimated the posterior means and variances of the user embeddings using the VAEs from the implicit data (e.g. click data). Item-specific weights of a MLP network are learned to map the user embeddings non-linearly into the multinomial predictions for the implicit data. Its extension~\citep{shenbin2020recvae} leverages users' embedding estimates from the previous iteration to construct a composite prior (with a standard Normal prior) for the user embeddings in the current iteration. This work also proposes to use a user-specific rescaling factor to control the KL-divergence for each user. However, to our best knowledge, there has not been any work which estimates the posterior parameters for both the user and the item embeddings using the VAEs.

\subsection{Other Neural Network Based Matrix Factorization Models}
Apart from the VAEs, there have been many other types of neural networks (NN) applied to the matrix factorization for collaborative filtering. \citet{liangvae2018} also proposed a denoising autoencoder (DAE) based user-oriented MF model. It randomly adds noise to the input data and then learns user embeddings to recover the original data. This model is very similar to another major DAE based user-oriented model proposed by~\citet{wu2016collaborative}. The main difference is that the former model adopts the multinomial log-likelihood loss which is more robust than the logistic loss employed by the latter. Earlier work has mostly used vanilla AEs to replace the MF approaches. One of the pioneering work~\citep{sedhain2015autorec} proposed two variants of the vanilla AEs for separately encoding users and items from the user-item and the item-user rating matrices respectively. Their decoders then reconstruct the above rating matrices from the resulting user and item embeddings respectively.

Unlike AE based approaches which take in the rating data, there are other NN based approaches that take in the one-hot encodings of the user and item IDs. \citet{he2017neural} proposed to capture non-linear interactions between the user and the item embeddings (mapped from their encoded IDs) via a MLP network. The two embeddings are concatenated before passed through the MLP. Different from the concatenation, \citet{he2018outer} proposed to perform outer product between the user and the item embeddings to better capture their interactive patterns. The resulting latent interaction maps are passed through a convolutional neural network (CNN) which learns high-order correlations among the embedding dimensions.

\citet{wang2019neural} proposed a graph neural network which leverages the message passing and aggregation of the node embeddings of items rated by a user for estimating his/her node embedding, and vice versa. This message-passing structure corresponds to the cross-feedback component of our framework. Despite this, our framework is fundamentally different from theirs as a fully probabilistic approach which learns a joint distribution with VAE based on a principled evidence lower bound.

\section{Proposed Framework}
In this paper, we use the symbols $I$ and $J$ to respectively denote the number of users and items involved in collaborative filtering. We denote the embedding of the $i$-th user as $\boldsymbol{u}_i$ and that of the $j$-th item as $\boldsymbol{v}_j$ where $i=1,...,I$ and $j=1,...,J$. The number of dimensions of $\boldsymbol{u}_i$ and $\boldsymbol{v}_j$ are $K$. The user and item embedding matrices are expressed as $\boldsymbol{U} \in \mathbbm{R}^{I\times K}$ and $\boldsymbol{V} \in \mathbbm{R}^{J\times K}$ respectively. The ratings of all the users on all the items are formulated as a matrix $\boldsymbol{X} \in \mathbbm{R}^{I\times J}$. Its ($i$,$j$)-th entry stores the rating $x_{ij}$ given by the $i$-th user to the $j$-th item. If the rating is missing, we set $x_{ij}=0$ for convenience of mathematical expression.  

%Our framework is fit to minimize the mean total difference between each observed entry $r_{ij}$ and its prediction $\hat{r}_{ij}$. In collaborative filtering, $\hat{r}_{ij}$ is computed as the dot product between $\boldsymbol{u}_i$ and $\boldsymbol{v}_j$, that is $\boldsymbol{u}_i\boldsymbol{v}^{T}_j=\sum^{K}_{k=1}u_{ik}v_{jk}$. Alternatively, the difference can be described with the matrix form $(\boldsymbol{R}-\boldsymbol{UV}^{T})\odot \boldsymbol{M}$. Here, $\boldsymbol{M} \in \mathbbm{R}^{I\times J}$ is a masking matrix. Its ($i$,$j$)-th entry is either zero if $r_{ij}=\text{``?''}$ or one otherwise.
\subsection{Deriving ELBOs for BMF Models}
Our framework is motivated by the derivation of two ELBOs for the BMF models in collaborative filtering which are user-oriented and item-oriented. More specifically, in the context of collaborative filtering, the ELBO specified in Eq.~\ref{eqn:elbo} is modified as follows:
\begin{equation}
\begin{split}
   Q=\mathbbm{E}_{q}\big[\log{p(\boldsymbol{X}, \boldsymbol{U},\boldsymbol{V})}\big] -\mathbbm{E}_{q}\big[\log{q(\boldsymbol{U},\boldsymbol{V})}\big]~\label{eqn:elbo_cf}
\end{split}
\end{equation}
From a \textit{user-oriented} viewpoint, Eq.~\ref{eqn:elbo_cf} can be further expanded as follows:
\begin{equation}
\begin{split}
   Q_{\boldsymbol{U}}= \sum\nolimits^{I}_{i}\Big( \mathbbm{E}[ \log p(\boldsymbol{u}_i | \boldsymbol{V}_i, \boldsymbol{X}_i) ] - \mathbbm{E}_{i} [ \log q(\boldsymbol{u}_i) ]\Big)+\text{C}_{1}~\label{eqn:elbo_cf1}
\end{split}
\end{equation}
%\begin{equation}
%\begin{split}
%   Q_j&=\mathbbm{E}_{q}\big[\log{p(\boldsymbol{X}, \boldsymbol{U},\boldsymbol{V})}\big] -\mathbbm{E}_{q}\big[\log{q(\boldsymbol{U},\boldsymbol{V})}\big]\hspace{55pt}\\&= \sum\nolimits^{J}_{j}\Big( \mathbbm{E}[ \log p(\boldsymbol{v}_j | \boldsymbol{U}_j, \boldsymbol{X}_j) ] - \mathbbm{E}_{j} [ \log q(\boldsymbol{v}_j) ]\Big)~\label{eqn:elbo1}
%\end{split}
%\end{equation}
Eq.~\ref{eqn:elbo_cf1} is derived under the mean-field assumption. The symbols $\boldsymbol{V}_i$ and $\boldsymbol{X}_i$ correspond to the embeddings of items rated by the $i$-th user and their received ratings. For $\boldsymbol{V}_i$, they are typically obtained from the one-hot encoding vector of the items' IDs via an embedding layer~\citep{sedhain2015autorec,he2017neural} and are fixed when Eq.~\ref{eqn:elbo_cf1} is maximized with respect to $q(\boldsymbol{u}_i)$. $p(\boldsymbol{u}_i | \boldsymbol{V}_i, \boldsymbol{X}_i)$ is the true posterior distribution to be approximated by the variational distribution $q(\boldsymbol{u}_i)$; that is to construct $q(\boldsymbol{u}_i)$ with the same form of $p(\boldsymbol{u}_i | \boldsymbol{V}_i, \boldsymbol{X}_i)$ and make it close enough with respect to its parameters. Terms that do not depend on $q(\boldsymbol{u}_i)$ are treated as the constant $\text{C}_1$. Likewise, Eq.~\ref{eqn:elbo_cf} can be expanded from an \textit{item-oriented} viewpoint as well: 
\begin{equation}
\begin{split}
   Q_{\boldsymbol{V}}= \sum\nolimits^{J}_{j}\Big( \mathbbm{E}[ \log p(\boldsymbol{v}_j | \boldsymbol{U}_j, \boldsymbol{X}_j) ] - \mathbbm{E}_{j} [ \log q(\boldsymbol{v}_j) ]\Big)+\text{C}_{2}~\label{eqn:elbo_cf2}
\end{split}
\end{equation}
Eq.~\ref{eqn:elbo_cf2} focuses on approximation with respect to $q(\boldsymbol{v}_j)$. All its terms have the same meanings as their counterparts in Eq.~\ref{eqn:elbo_cf1}. For $\boldsymbol{U}_j$, they are usually derived from the users' IDs via another embedding layer and are fixed when Eq.~\ref{eqn:elbo_cf2} is maximized with respect to $q(\boldsymbol{v}_j)$. The true posterior distribution $p(\boldsymbol{v}_j | \boldsymbol{U}_i, \boldsymbol{X}_j)$, in this case, is approximated by the variational distribution $q(\boldsymbol{v}_j)$.

From Eq.~\ref{eqn:elbo_cf1} and Eq.~\ref{eqn:elbo_cf2}, we see that neither user-oriented nor item-oriented viewpoints capture the full inter-dependency among users, items and ratings. To do so, we consider a unified viewpoint to expand Eq.~\ref{eqn:elbo_cf}; that is to construct $q(\boldsymbol{u}_i)$ and $q(\boldsymbol{v}_j)$ to depend on both the data and each other for KL-divergence:
\begin{equation}
\begin{split}
    Q=\mathbbm{E}\big[\log{p(\boldsymbol{X}|\boldsymbol{U},\boldsymbol{V})}\big]&-\beta_{\boldsymbol{u}} \sum\nolimits^{I}_{i}\text{KL}\big[q(\boldsymbol{u}_i|\boldsymbol{V}_i, \boldsymbol{X}_i)||p(\boldsymbol{u}_i)\big]\\&- \beta_{\boldsymbol{v}}\sum\nolimits^{J}_{j}\text{KL}\big[q(\boldsymbol{v}_j|\boldsymbol{U}_j, \boldsymbol{X}_j)||p(\boldsymbol{v}_j)\big]~\label{eqn:elbo_cf3}
\end{split}
\end{equation}
Eq.~\ref{eqn:elbo_cf3} can be interpreted as an autoencoder. It ``encodes'' data $\boldsymbol{X}_i$ and variables  $\boldsymbol{V}_i$ into $\boldsymbol{u}_i$, and meanwhile, data $\boldsymbol{X}_j$ and variables  $\boldsymbol{U}_j$ into $\boldsymbol{v}_i$. Then, it ``decodes'' the user and item variables to reconstruct the data. $\beta_{\boldsymbol{u}}$ and $\beta_{\boldsymbol{v}}$ denote the hyper-parameters that respectively control the regularization effect of the user-specific and item-specific KL terms according to~\citep{liangvae2018}.
%herefore, Eq.s~\ref{eqn:elbo1} and~\ref{eqn:elbo2} motivate us to alternately estimate $\boldsymbol{U}$ and $\boldsymbol{V}$ with one conditioned on the other via a cross-feedback connectivity which will be specified in the next section. 
%Our framework focuses on inferring the posterior distributions for the user and item embeddings. It leverages the mean-field variational inference method. This method assumes that the embeddings are independent and estimates their posterior parameters by minimizing the negative evidence lower bound (ELBO) loss. 

%In collaborative filtering, the embeddings are usually assumed to follow Normal distributions. In this case, the negative ELBO loss, denoted by $\text{Q}$, for the Normal likelihood over the ratings $\boldsymbol{R}$ can be expressed as follows:

\subsection{Framework Architecture}
To optimize the ELBO specified by Eq.~\ref{eqn:elbo_cf3}, we propose an VAE-based BMF framework for collaborative filtering. This framework is characterized by a cross-feedback connectivity component which captures the mutual dependency between user and item variables in the equation.
\begin{figure}[t]
%\begin{subfigure}[b]{0.7\columnwidth}
\centering
\includegraphics[width=0.8\textwidth]{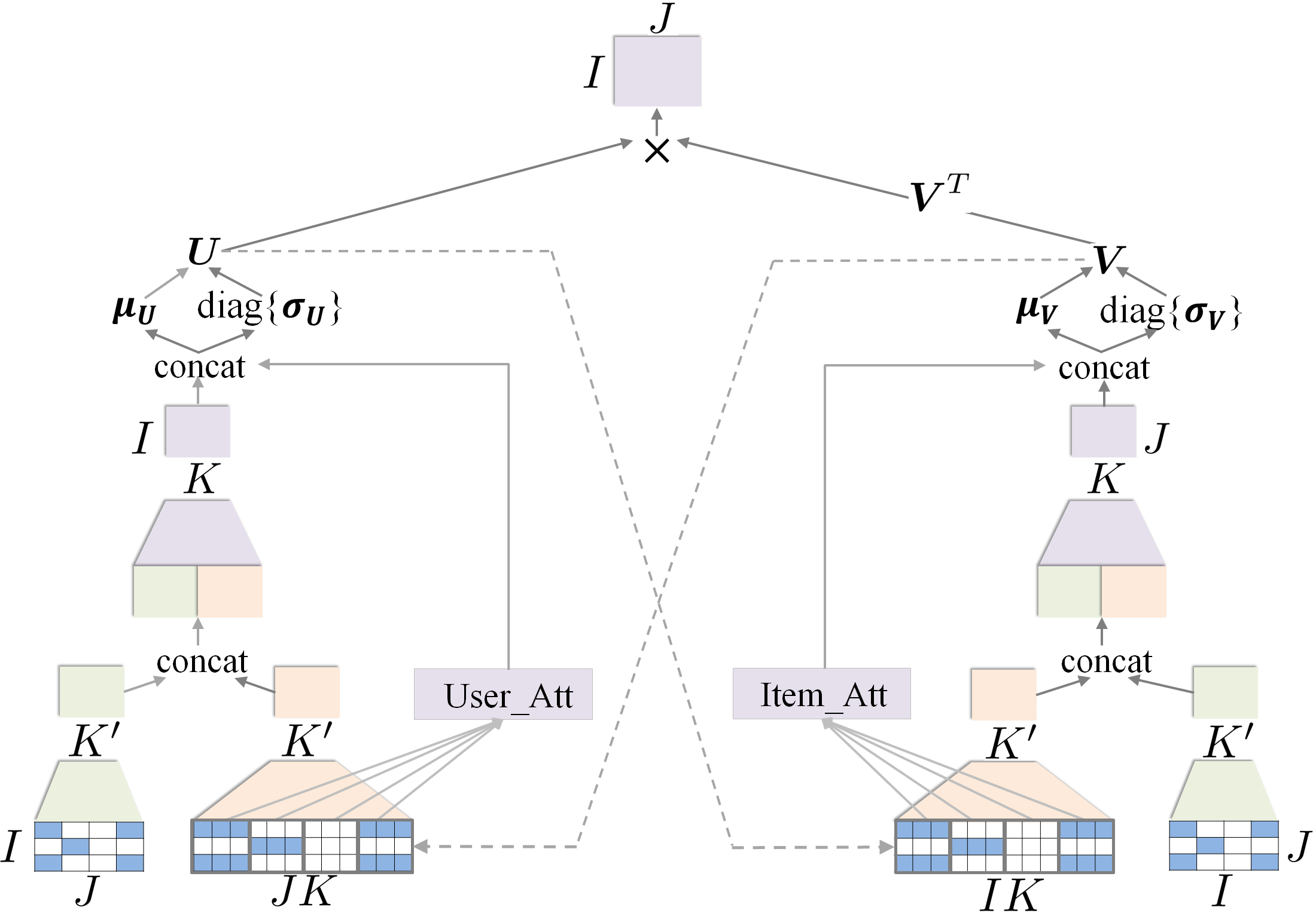}
  %\caption{Ability-Difficulty-Response Interaction}  
\caption{The architecture of our framework.}
%  \end{subfigure}
%  \begin{subfigure}[b]{0.7\columnwidth}
%  \centering
%\includegraphics[width=2in]{images/encoder.png}
  %\caption{Ability-Difficulty-Response Interaction}  
%  \caption{}
%  \end{subfigure}
%    \begin{subfigure}[b]{0.7\columnwidth}
%  \centering
%\includegraphics[width=1.5in]{images/decoder_model.png}
  %\caption{Ability-Difficulty-Response Interaction}  
%  \caption{}
%  \end{subfigure}
\label{fig:vae_bmf}
\end{figure}
Figure~\ref{fig:vae_bmf} shows the architecture of our framework. Note that for simplicity, the posterior means and variances are shown to be computed by the same set of feed-forward networks. In the experiments, we implement separate MLP feed-forward networks to compute these two sets of parameters.

Our framework inherits the \textit{encoder-decoder} structure from the VAEs. In terms of the decoder, it reconstructs the rating matrix $\boldsymbol{X}$ as the multiplication of the user and item embedding matrices $\boldsymbol{UV}^{T}$. In this paper, we focus on reconstructing the observed data during the training phase. Therefore, the log-likelihood term $p(\boldsymbol{X}|\boldsymbol{U},\boldsymbol{V})$ in Eq.~\ref{eqn:elbo_cf3} is calculated solely based on each observed entry of $\boldsymbol{X}$ (i.e. $x_{ij}\neq0$). As for the encoder part, there are user-side and item-side encoders which respectively model the variational distributions $q(\boldsymbol{u}_i|\boldsymbol{V}_i, \boldsymbol{X}_i)$ and $q(\boldsymbol{v}_j|\boldsymbol{U}_j, \boldsymbol{X}_j)$ to generate the matrices $\boldsymbol{U}$ and $\boldsymbol{V}$. Both encoders possess two types of inputs: the \textit{observed rating} input and the \textit{latent embedding} input. The user-side observed input (layer) takes in the matrix $\boldsymbol{X}$, while the item-side input receives its transpose $\boldsymbol{X}^{T}$. 

As for the latent input (layer) of the user-side encoder, it is constructed based on the item embeddings. More specifically, it is constructed as a matrix $\boldsymbol{Z}_{\boldsymbol{U}} \in \mathbbm{R}^{I\times (JK)}$. Its $i$-th row is a concatenation of the embeddings of all the items after the embeddings of those unrated by the $i$-th user are masked by zeros. As an example, suppose the $i$-th user has rated the first of three items and the embedding dimension $K=2$. Then, the input row corresponding to this user stores a vector $\boldsymbol{z}_{\boldsymbol{u}_i} = [v_{11},v_{12},0,0,0,0]$. Likewise, the latent input layer of the item-side encoder takes in a matrix $\boldsymbol{Z}_{\boldsymbol{V}}\in \mathbbm{R}^{J\times (IK)}$ which is constructed in the same way as $\boldsymbol{Z}_{\boldsymbol{U}}$.

Both the observed and latent inputs at each side are passed through and transformed by their own MLP networks to compute the corresponding posterior parameters. To make the framework description succinct, we only describe the MLP networks of the user-side encoder that compute the posterior means for the user embeddings. Initially, the rating input of the user-side encoder is transformed by an MLP network as follows:
\begin{eqnarray}
\begin{split}
\boldsymbol{H}_{1}=\text{ReLU}(\boldsymbol{X}\boldsymbol{W}_{0}+\boldsymbol{{1}}\boldsymbol{\alpha}_{0})\\
\boldsymbol{H}_{2}=\text{ReLU}(\boldsymbol{H}_{1}\boldsymbol{W}_{1}+\boldsymbol{{1}}\boldsymbol{\alpha}_{1})\\
\ldots\hspace{50pt}\\
\boldsymbol{H}_{L'+1}=\text{ReLU}(\boldsymbol{H}_{L'}\boldsymbol{W}_{L'}+\boldsymbol{{1}}\boldsymbol{\alpha}_{L'})\\
\end{split}~\label{eqn:recursive_fun}
\end{eqnarray}
In Eq.~\ref{eqn:recursive_fun}, $\boldsymbol{W}_{0} \in \mathbbm{R}^{J\times M^{(1)}}$ and $\boldsymbol{\alpha}_{0} \in \mathbbm{R}^{1\times M^{(1)}}$ are respectively the weight matrix and the bias vector specific to the observed input layer. $M^{(1)}$ is the number of neurons for the first hidden layer subsequent to the observed inputs. The term $\boldsymbol{{1}}$ denotes a column vector of size $I\times 1$ with its entries being all ones. The matrix output of this input layer is then transformed by the activation function ReLU. The result $\boldsymbol{H}_{1} \in \mathbbm{R}^{I\times M^{(1)}}$ serves as the input to the first hidden layer. There are in total $L'$ hidden layers in the user-side encoder for the observed inputs. The output of the $L'$-th hidden layer, $\boldsymbol{H}_{L'+1} \in \mathbbm{R}^{I\times M^{(L'+1)}}$, is the intermediate user embedding matrices derived from the rating input $\boldsymbol{X}$. For convenience, we use $K'=M^{(L'+1)}$ to denote the dimension of each intermediate user embedding.

Similar to the observed input layer, the latent embedding input $\boldsymbol{Z}_{\boldsymbol{U}}$ is passed through its dedicated MLP network. Correspondingly, we denote the weight matrices and the bias vectors of each hidden layer of this MLP network as $\{\boldsymbol{\Phi}_{l'},\boldsymbol{\beta}_{l'}\}_{l'=1,...,L'}$. Note that the weight matrix of the latent input layer $\boldsymbol{\Phi}_{0} \in \mathbbm{R}^{JK\times M^{(1)}}$ is much larger than its counterpart $\boldsymbol{W}_{0}$ for the observed input layer. Other than this, we let the two MLP encoders have the same number of hidden layers $L'$, the same number of hidden neurons $M^{(l')}$ at the $l'$-th layer, the same setting of the activation functions and the same dimension $K'$ for the intermediate embeddings\footnote{In the experiment, we did not find notable improvements in the framework performance by making the two MLP networks have different numbers for these parameters.}. Finally, we denote the output of the $L'$-th hidden layer over the latent inputs as $\boldsymbol{G}_{L'+1}$. 

After generating the intermediate embedding matrices $\boldsymbol{H}_{L'+1}$ and $\boldsymbol{G}_{L'+1}$, we concatenate them in a column-wise manner. This creates the intermediate input (layer) $\boldsymbol{S}=[\boldsymbol{H}_{L'+1};~\boldsymbol{G}_{L'+1}] \in \mathbbm{R}^{I\times 2K'}$ to the final MLP network. It computes the posterior means with another set of weight matrics and bias vectors: $\{\boldsymbol{\Psi}_{l},\boldsymbol{\gamma}_{l}\}_{l=1,...,L}$. This network can have different numbers of hidden layers $L$ and neurons $M^{(l)}$ at each layer from the previous networks. Furthermore, the linear function is used for activation of its output layer to compute the posterior means $\{\boldsymbol{\mu}_i\}_{i=1,...,I}$.

The posterior variances $\{\text{diag}(\boldsymbol{\sigma}^2_{i})\}_{i=1,...,I}$ is computed in the same way as the posterior means but with different sets of weights and biases and with the acitivation of the MLP output being the Sigmoid function. Finally, each user embedding $\boldsymbol{u}_i$ in the matrix $\boldsymbol{U}$ is generated based on the corresponding  $\boldsymbol{\mu}_i$ and $\text{diag}(\boldsymbol{\sigma}^2_{i})$ using the reparametrization trick~\citep{kingma2013auto}.  In addition, we have also tried to clip the embeddings to make them non-negative. However, in the experiment, we have not found any difference in performance resulted from the clipping operation. 

\subsection{Cross-Feedback Connectivity}
The cross-feedback connectivity provides a solution to effectively modelling the variational distributions $q(\boldsymbol{u}_i|\boldsymbol{V}_i, \boldsymbol{X}_i)$ and $q(\boldsymbol{v}_j|\boldsymbol{U}_j, \boldsymbol{X}_j)$ in Eq.~\ref{eqn:elbo_cf3}. With such connectivity, we are able to optimize an ELBO that aims to capture the full inter-dependency among users, items and ratings. More specifically, in our framework, there exist two cross-feedback loops. As shown by Figure~\ref{fig:vae_bmf}, one feeds the user-side (encoder) output back into the item-side latent input layer. The other feeds the item-side output back into the user-side latent input layer. According to Algorithm 1, this cross feedback mechanism is conducted iteratively alongside the variational inference for the embeddings $\boldsymbol{U}$ and $\boldsymbol{V}$. In our algorithm, steps 6 and 15 specify the cross feedback loops between the latent inputs at one side and the encoder outputs at the other side.

In Algorithm 1, users and items are batched according to the predefined user and item batch sizes $B_{\boldsymbol{U}}$ and $B_{\boldsymbol{V}}$. As a result, the observed and latent inputs of the encoders at both sides need to be sliced by the corresponding batches before passed through the networks. In addition, our algorithm employs nested iteration across the item and the user batches rather than sequential iteration. The latter alternates the update of the user-side and the item-side weights and biases while the nested iteration updates these parameters all at once. We found in the experiment that the nested iteration allows our framework to achieve higher prediction accuracy at the cost of more running time.

  \begin{algorithm}[t]
   \caption{VAE-BMF with Cross-Feedback Connectivity}
    \textbf{Input:} ratings $\boldsymbol{X}$, user batch size $B_{\boldsymbol{U}}$, item batch size $B_{\boldsymbol{V}}$\\
    Initialize entries of $\boldsymbol{U}, \boldsymbol{V}$ by sampling them from $\mathcal{N}(\mu, \sigma^2)$\\
    Sample $\left \lceil{\frac{I}{ B_{\boldsymbol{U}} }}\right \rceil$ user batches and $\left \lceil{\frac{I}{ B_{\boldsymbol{V}} }}\right \rceil$ item batches\\ 
    Randomly initialize user and item-side encoder parameters\\
    \For{not converged or not max iterations}{
    Build user-side latent inputs $\boldsymbol{Z}_{\boldsymbol{U}}$ from $\boldsymbol{V}$ and item-side latent inputs $\boldsymbol{Z}_{\boldsymbol{V}}$ from $\boldsymbol{U}$\\
    \For{each user batch}{
    Slice $\boldsymbol{X},\boldsymbol{Z}_{\boldsymbol{U}}$ by the user batch\\
    \For{each item batch}{
    Slice $\boldsymbol{X}^{T},\boldsymbol{Z}_{\boldsymbol{V}}$ by the item batch\\
    Compute gradients $\nabla\text{Q}$ specified in Eq.~\ref{eqn:elbo_cf3} w.r.t. all the weights and biases over the slices\\
    Update the weights and biases with the computed gradients 
    }
    }
    Resample $\boldsymbol{U}, \boldsymbol{V}$ using reparametrization trick
    }
  \end{algorithm}

\subsection{User-Side and Item-Side Attention}
To further strengthen the effect of the cross-feedback connectivity, we propose a user-side and item-side attention component based on the latent inputs of each side. The basic idea is that some items tend to be more representative of a user's interest or taste, while some users' tastes might better represent the quality or characteristics of an item. Furthermore, the encoders of our framework transform the entire latent inputs $\boldsymbol{Z}_{\boldsymbol{U}}$ and $\boldsymbol{Z}_{\boldsymbol{V}}$ without recognizing the individual embeddings underneath and their relationships with the target embeddings. In contrast, the two attention mechanisms capture such relationships for the user and item embeddings respectively. 

To make a succinct description, only the user-side attention is specified here while the item-side attention, which has the same format and a counterpart set of parameters, is omitted. In this case, the posterior mean vector $\boldsymbol{\mu}_i$ of the $i$-th user is attended to the embedding of the $j$-th item with a weight $a_{ij}$ calculated as follows:
\begin{equation}
    a_{ij}=\frac{\mathbbm{1}_{\{\boldsymbol{v}_{j}\in\boldsymbol{V}_{i}\}}\boldsymbol{\mu}_i\boldsymbol{\Theta}\boldsymbol{v}^{T}_j}{\sum\nolimits^{J}_{j'=1}\mathbbm{1}_{\{\boldsymbol{v}_{j'}\in\boldsymbol{V}_{i}\}}\boldsymbol{\mu}_i\boldsymbol{\Theta}\boldsymbol{v}^{T}_{j'}}\label{eqn:att_weight}
\end{equation}
where $\boldsymbol{\Theta}\in \mathbbm{R}^{K\times K}$ are the weight matrix to be learned at the user side. Then, we compute the weighted average of the item embeddings according to their weights to obtain the attention vector $\boldsymbol{c}_i$ for the $i$-th user:
\begin{equation}
    \boldsymbol{c}_i=\sum\nolimits^{J}_{j=1}\mathbbm{1}_{\{\boldsymbol{v}_{j}\in\boldsymbol{V}_{i}\}}a_{ij}\boldsymbol{v}_j\label{eqn:att_vec}
\end{equation}
Note that in Eq.~\ref{eqn:att_weight} and Eq.~\ref{eqn:att_vec}, only the items rated by the $i$-th user (as shown by the indicator functions) are attended to by $\boldsymbol{\mu}_i$. We find in the experiment that this attention strategy improves the performance of our framework when responses are abundant but harm the performance when responses are sparse. In the latter case, we instead attend $\boldsymbol{\mu}_i$ to all the item embeddings. By doing so, the attention of a user embedding moves from being local to being global by extracting useful information from the entire item population. Finally, the attention vector $\boldsymbol{c}_i$ can be integrated with the original vector $\boldsymbol{\mu}_i$ in different ways to form a new user embedding, such as element-wise addition: $\boldsymbol{\mu}_i:= \boldsymbol{\mu}_i + \boldsymbol{c}_i$; or concatenation: $\boldsymbol{\mu}_i:= [\boldsymbol{\mu}_i; \boldsymbol{c}_i]$. In this paper, we choose to use the concatenation strategy. The new user embedding $\boldsymbol{\mu}_i$ thus has a dimension of $2K$ and we transform it back to be $K$-dimensional by simply placing a linear layer on top of it which we omit in Figure~\ref{fig:vae_bmf} for the simplicity of presentation. Likewise, the variance vector $\text{diag}(\boldsymbol{\sigma}^2_i)$ for the $i$-th user is also attended to the item embeddings by following the same strategy with a different attention weight matrix $\boldsymbol{\Lambda}$. 

\section{Experiments and Results}
In this section, we compare our framework with various state-of-the-art CF models across five real-world datasets. Furthermore, we study the importance of the cross-feedback and attention mechanisms in improving the model performance and accelerating the model convergence. This includes an ablation study and a convergence analysis in which we compare the performance and the convergence (rate) of different variants of the framework by taking away its different components.

\subsection{Datasets}
We investigate five medium to large real-world datasets. The major information of these datasets is summarized in Table \ref{tab:datasets}.

\begin{table}[h]
    \centering
    \scriptsize
    \begin{tabular}{lccc}
    \toprule
        Datasets & \# users & \# items & \# ratings\\
    \hline
         ML-1M & 6,040  & 3,900  & 1,000,209 \\
         ML-10M & 69,878  & 10,631 & 9,999,989  \\
         ML-20M & 135,543 &  21,651  & 19,972,967  \\
         Amazon Pet & 46,403  & 61,673  & 1,008,336   \\
         Amazon Kindle & 92,351  & 55,924  &  1,619,155  \\
    \bottomrule
    \end{tabular}
    \caption{The summary of the key attributes of the datasets used in the experiment.}
    \label{tab:datasets}
\end{table}

\textbf{MovieLens 1M, 10M, 20M} (ML-1M, ML-10M, ML-20M): These are medium to large datasets of user-movie ratings collected from MovieLens\footnote{https://movielens.org/}, a movie recommendation service, over different periods of time.

\textbf{Amazon pet supplies} and \textbf{Kindle stores} (Amazon-Pet, Amazon-Kindle): These datasets contain ratings from Amazon on different types of products\footnote{http://jmcauley.ucsd.edu/data/amazon/}. For these two datasets, we only preserve users and items that have at least 10 ratings.

\subsection{Experimental setup}
To evaluate the performance of our framework, we randomly split the observed rating data into the training, validation and testing data subsets with a 70\%-15\%-15\% ratio. The evaluation metrics employed in the experiment include the RMSE, Recall@\textit{N} and NDCG@\textit{N} where \textit{N} = 20 or 50. The RMSE measures the difference between the observed and predicted ratings irrespective of the specific users or items. The Recall and NDCG metrics are measured specific to each user and are originally designed for binary ratings (e.g. clicks). In this case, we binarize the ratings accordingly (i.e. 1 if the rating is greater than 3, suggesting that a user is interested; 0 otherwise, for not being interested). The difference between the Recall and the NDCG metrics is that the former accounts for all the top-\textit{N} ranked items as being equally important, while the latter assigns more weights to items ranked higher. Here, we calculate the average Recall and NDCG scores across all the users who rated at least two items by following the formulas specified in~\citep{liangvae2018}. Note that in~\citep{liangvae2018}, test datasets were held out specific to users while in this work, they are held out regardless of users and items.

We perform hyper-parameter optimization for our framework based on its RMSE scores on the validation datesets. More specifically, a grid search is conducted on the embedding dimensions $K$ and $K'$ among $\{10,15,25,50\}$, the number of hidden layers $L$ and $L'$ among $\{0,1,2\}$, the number of hidden neurons $M^{(l)}$ and $M^{(l')}$ at each layer among $\{10,25,50,100\}$ and the controlling parameters $\beta_{\boldsymbol{u}},\beta_{\boldsymbol{v}}$ for the KL terms among $\{10^{-5}, 10^{-4},...,0.1, 1\}$. We restrict the grid search such that $K \leq K'$ and $K, K' \leq M^{(l)}, M^{(l')}$ at any hidden layer. As for the user and item batch sizes $B_{\boldsymbol{U}}$ and $B_{\boldsymbol{V}}$, we enlarge them when the numbers of users and items in the datasets are larger. This accelerates the framework training at the cost of more memory consumption and the degraded framework performance in RMSE. In the experiment, we carefully tuned the batch sizes considering this trade-off. For the ML-1M dataset, we set $B_{\boldsymbol{U}}, B_{\boldsymbol{V}}$ to be 100 given its relatively small numbers of users and items, while for the other datasets, $B_{\boldsymbol{U}}, B_{\boldsymbol{V}}$ are set to be 1,000. 

\subsection{Baselines}
We compare our framework with the following baselines.
\begin{itemize}
    \item \textbf{Neural Collaborative Filtering} (NCF)~\citep{he2017neural}: This model uses an MLP network (with embedding layers for user and item ID one-hot encoding vectors) to replace the traditional matrix factorization over the user and item embeddings. We adhere to the choice of ReLU as the activation function from the original paper.
    \item \textbf{Convolutional Neural Network based Collaborative Filtering} (CNN-CF)~\citep{he2018outer}: This model applies a CNN network to the outer product of the user and the item embeddings to capture the user-item interactive patterns. We adopt the default kernel size $2\times 2$ and the ReLU activation function from the original paper. Furthermore, we change the original Bayesian personalized ranking loss on the implicit data to the MSE loss on our explicit rating data.
    \item \textbf{Autoencoder based User-side \& Item-side Collaborative Filtering} (U-Rec, I-Rec)~\citep{sedhain2015autorec}: U-Rec encodes and decodes the rating matrix $\boldsymbol{X}$ with a standard autoencoder. The encoder in this case can only generate user embeddings. Likewise, I-Rec encodes and decodes $\boldsymbol{X}^{T}$ with the encoder only generating item embeddings. We adopt the ReLU function for the hidden layers and the Sigmoid function for the output layer as described in the paper.
    \item \textbf{Variational \& Denoising Autoencoder based Collaborative Filtering} (VAE-CF, DAE-CF)~\citep{liangvae2018}: These models are essentially user-side autoencoders that only generate embeddings for the users. We change their multinomial likelihood loss functions on the implicit data to the MSE loss over our rating data. The other parts of the models remain unchanged. These include the controlling parameter $\beta$ for the KL term and its decay strategy in VAE-CF, the weight decay strategies and the dropout probabilities in both models.
\end{itemize}

The optimization of the hyper-parameters for each of these models is based on the validation RMSE. Table~\ref{tab:hyperparams} summarizes the ranges of the hyper-parameter values of each model for the optimization. All the models, except for the VAE-CF, are regularized with $\text{L}_{2}$ norms on their parameters for which the controlling hyper-parameters are selected among $\{0,10^{-2},10^{-1},1\}$. For the VAE-CF model, the controlling parameter $\beta$ for its KL term is selected among $\{10^{-5}, 10^{-4},...,10^{-1},1\}$.

\begin{table}[t]
    \centering
    \scriptsize
    \begin{tabular}{lccccc}
    \toprule
        Models & $K$ & \# hidden layers & \# neurons &$\text{L}_2$\\
    \hline
        NCF&$\{5,10,25,50,100\}$&0 to 2&$\{10,25,50,100\}$&\{0,0.01,0.1,1\}\\
        CNN-CF&$\{8, 16, 32, 64, 128\}$&1 to 6&$\{64\times64,32\times32,...\}$&\{0,0.01,0.1,1\}\\
        U/I-Rec&$\{5,10,25,50,100\}$&0 to 2&$\{10,25,50,100\}$&\{0,0.01,0.1,1\}\\
        VAE-CF&$\{10,25,50,100,200\}$&0 to 2&$\{50, 100, 200, 400\}$&N/A\\
        DAE-CF&$\{10,25,50,100,200\}$&0 to 2&$\{50, 100, 200, 400\}$&\{0,0.01,0.1,1\}\\
        
    \bottomrule
    \end{tabular}
    \caption{The hyper-parameter value ranges of each baseline model for the hyper-parameter optimization on the validation RMSE.}
    \label{tab:hyperparams}
\end{table}

\begin{table}[t]
\centering
\scriptsize
  \begin{tabular}{llcccccc}
    \toprule
     Metrics&Models& ML-1M & ML-10M & ML-20M & Amazon-Pet & Amazon-Kindle\\
    \midrule
    &\textbf{NCF} & 0.870  & 0.831  & 0.822  & \underline{1.098}  &\underline{0.734}\\
        &\textbf{CNN-CF} & 0.875 & 0.844  & 0.839  & 1.119  &0.770\\
    &\textbf{U-Rec} & 0.883  & 0.893 & 0.833  & 1.136  &0.772\\
    &\textbf{I-Rec} & \underline{0.859}  & \underline{0.823} & \underline{0.818}  & 1.213  &0.787\\
    RMSE&\textbf{DAE-CF} & 0.911 &  0.852 & 0.841  & 1.247  & 0.762\\
    &\textbf{VAE-CF} & 0.903 & 0.834  & 0.836  &1.211  &0.747\\
    &\textbf{Ours} & \textbf{0.842} & \textbf{0.804} & \textbf{0.798}  & \textbf{1.074} & \textbf{0.702}\\
    \hline
    &\textbf{NCF} &0.797&0.799&0.797&0.908&\underline{0.963}\\
        &\textbf{CNN-CF}  &0.774&0.778&0.796&0.886&0.948\\
    &\textbf{U-Rec}  &0.796&0.794&0.802&0.896&\underline{0.963}\\
    &\textbf{I-Rec}  &0.767&0.759&0.776&0.883&0.942\\
    NDCG@20&\textbf{DAE-CF}  &0.797&\underline{0.801}&0.799&0.899&0.961\\
    &\textbf{VAE-CF}  &\underline{0.798}&\underline{0.801}&\underline{0.803}&\underline{0.909}&0.962\\
    &\textbf{Ours}  &\textbf{0.802}&\textbf{0.806}&\textbf{0.809}&\textbf{0.911}&\textbf{0.966}\\
        \hline
        &\textbf{NCF} &0.788&0.791&0.791&0.907&\underline{0.963}\\
        &\textbf{CNN-CF}  &0.758&0.762&0.790&0.882&0.946\\
    &\textbf{U-Rec}  &0.788&0.788&0.796&0.894&\underline{0.963}\\
    &\textbf{I-Rec}  &0.754&0.744&0.771&0.879&0.941\\
    NDCG@50&\textbf{DAE-CF}  &\underline{0.789}&0.793&\underline{0.795}&0.898&0.960\\
    &\textbf{VAE-CF}  &0.787&\underline{0.794}&0.793&\underline{0.908}&0.960\\
    &\textbf{Ours}  &\textbf{0.793}&\textbf{0.797}&\textbf{0.800}&\textbf{0.910}&\textbf{0.965}\\
        \hline
    &\textbf{NCF} &0.775&0.778&0.790&\underline{0.905}&\underline{0.959}\\
        &\textbf{CNN-CF}  &0.756&0.761&0.792&0.882&0.939\\
    &\textbf{U-Rec}  &0.776&0.781&0.784&0.893&0.958\\
    &\textbf{I-Rec}  &0.749&0.740&0.755&0.879&0.933\\
    Recall@20&\textbf{DAE-CF}  &\underline{0.778}&\underline{0.782}&0.785&0.896&0.958\\
    &\textbf{VAE-CF}  &0.777&\underline{0.782}&\underline{0.786}&\textbf{0.906}&0.958\\
    &\textbf{Ours}  &\textbf{0.781}&\textbf{0.784}&\textbf{0.790}&\textbf{0.906}&\textbf{0.961}\\
        \hline
        &\textbf{NCF} &0.763&0.768&\underline{0.782}&0.904&\underline{0.958}\\
        &\textbf{CNN-CF}  &0.742&0.749&0.779&0.880&0.937\\
    &\textbf{U-Rec}  &0.764&0.763&0.775&0.892&0.957\\
    &\textbf{I-Rec}  &0.738&0.726&0.743&0.876&0.932\\
    Recall@50&\textbf{DAE-CF}  &0.766&0.772&\underline{0.782}&0.896&0.957\\
    &\textbf{VAE-CF}  &\underline{0.767}&\underline{0.773}&\textbf{0.783}&\underline{0.905}&\underline{0.958}\\
    &\textbf{Ours}  &\textbf{0.770}&\textbf{0.774}&\textbf{0.783}&\textbf{0.906}&\textbf{0.961}\\
    \bottomrule
  \end{tabular}
  \caption{Evaluation results of each model across the five datasets with different metrics.}
  \label{tab:metric_results}
\end{table}

\subsection{Predictive Analysis}
In this section, we evaluate the effectiveness of our framework by comparing its performance against that of the baseline models on predicting the observed data from the testing datasets. Table~\ref{tab:metric_results} summarizes the testing RMSE, Recall@\textit{N} and NDCG@\textit{N}, where \textit{N} = 20 or 50, of each model over the five benchmark datasets. It can be observed that our framework has achieved lower testing RMSE than all the baselines on these datasets. More specifically, it manages to outperform the second-best model by at least 0.017 (on the ML-1M dataset) and at most 0.032 (on the Amazon-Kindle dataset). It is also interesting to see that the I-Rec model holds the second place across all the Movielens datasets while it is the NCF model for the Amazon datasets. We conjecture that this has to do with how the I-Rec model works. It performs better when ratings are abundant across the items which is the case in the Movielens data. However, we can see from Table~\ref{tab:datasets} that the ratings in the Amazon datasets are much sparser across the users and the items. The sparsity at both sides cause the I-Rec and the U-Rec to degrade in their performance. In comparison, our framework appears to be more robust when the sparsity occurs in the Amazon datasets. 

\begin{figure*}[t]
    \centering
    \begin{subfigure}[t]{0.45\textwidth}
        \centering
        \includegraphics[height=1.3in]{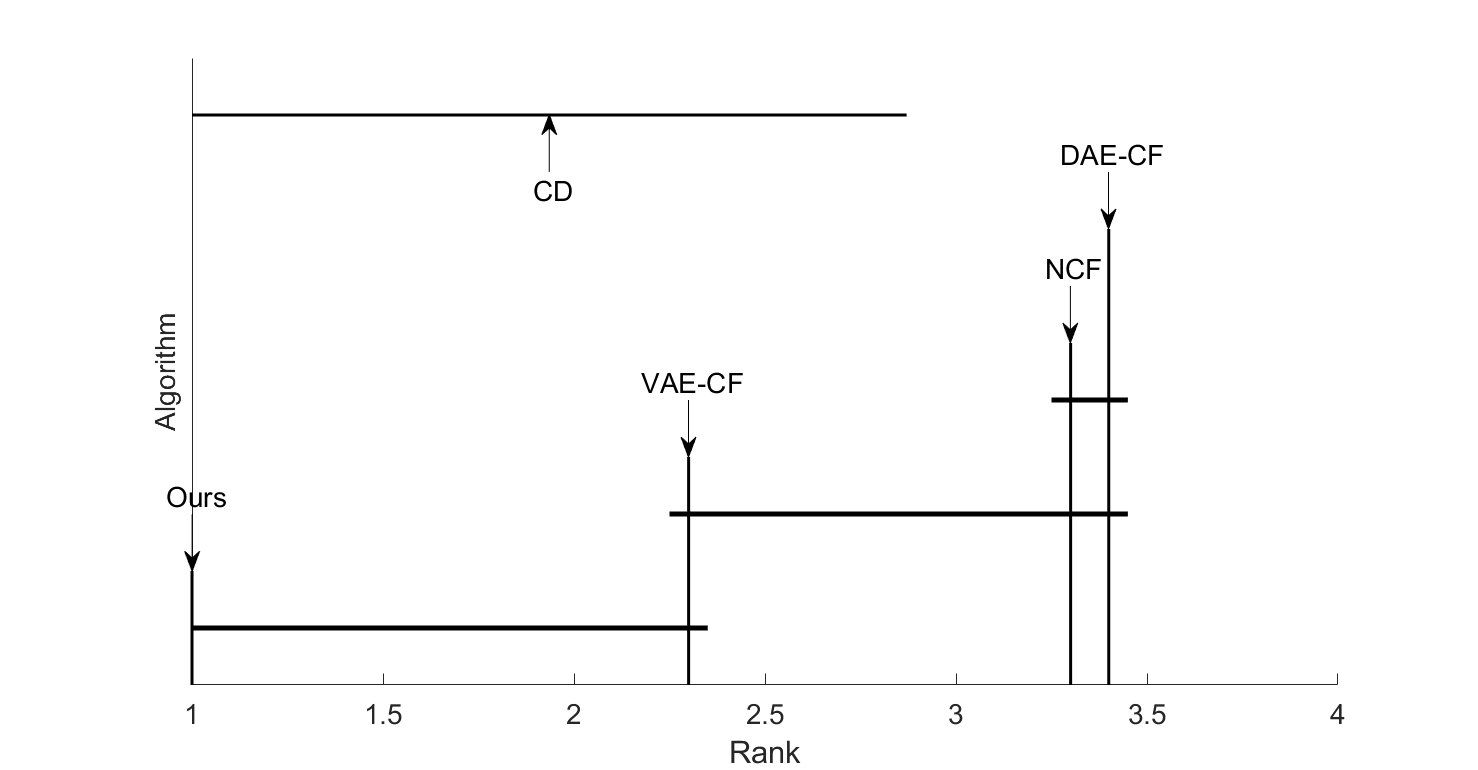}
        \caption{The CD diagram for NDCG@20}~\label{fig:ndcg20_cd}
    \end{subfigure}%
    \begin{subfigure}[t]{0.45\textwidth}
        \centering
        \includegraphics[height=1.3in]{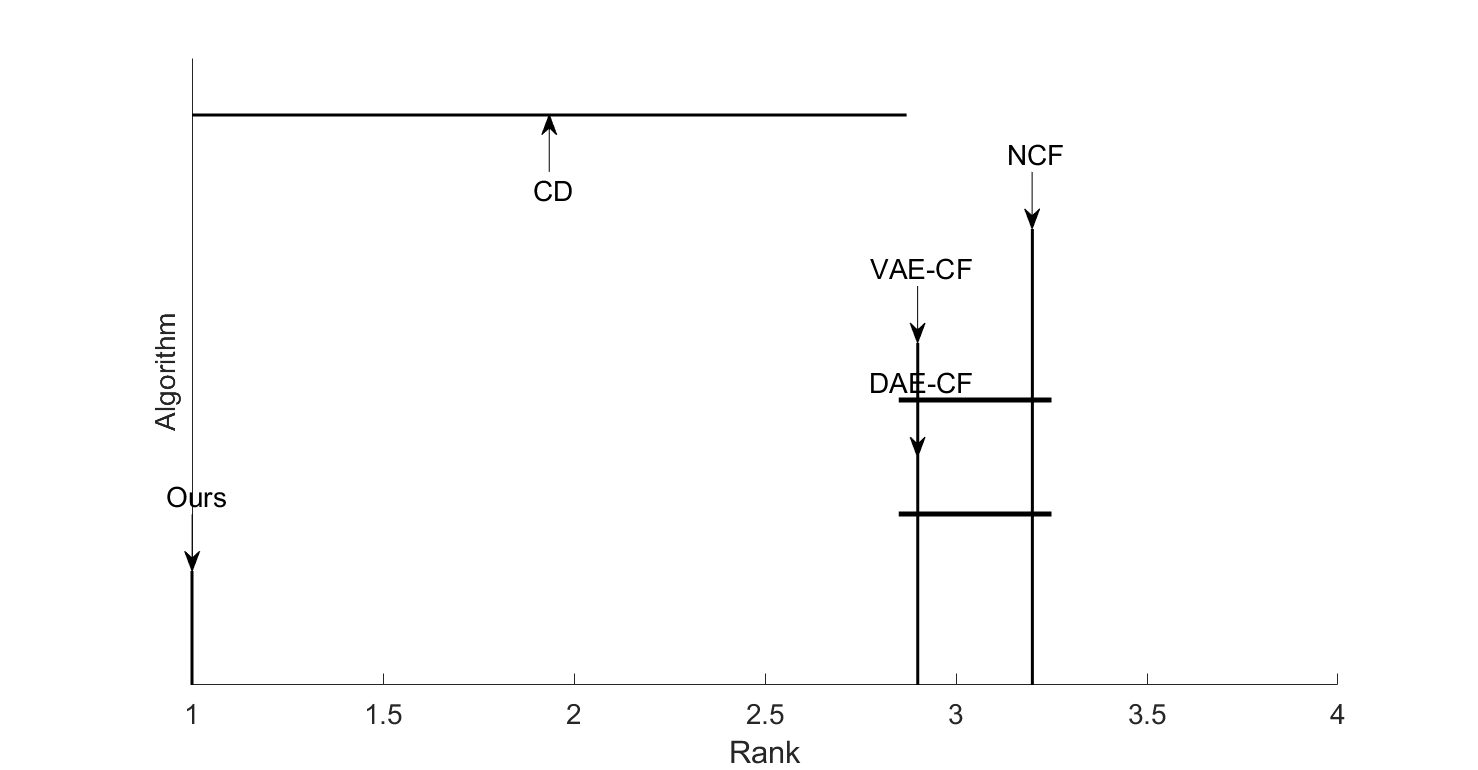}
        \caption{The CD diagram for NDCG@50}~\label{fig:ndcg50_cd}
    \end{subfigure}
        \begin{subfigure}[t]{0.45\textwidth}
        \centering
        \includegraphics[height=1.3in]{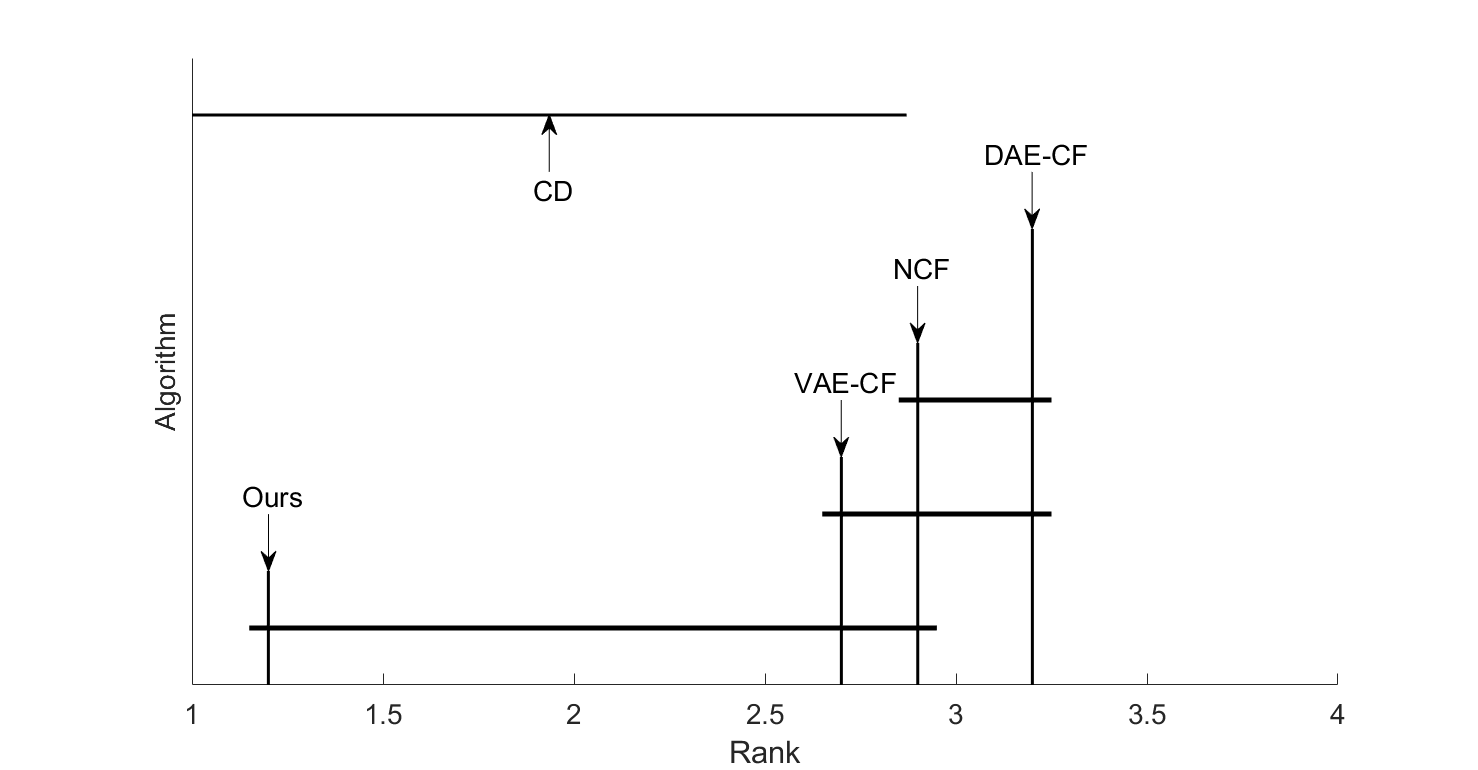}
        \caption{The CD diagram for Recall@20}~\label{fig:recall20_cd}
    \end{subfigure}
            \begin{subfigure}[t]{0.45\textwidth}
        \centering
        \includegraphics[height=1.3in]{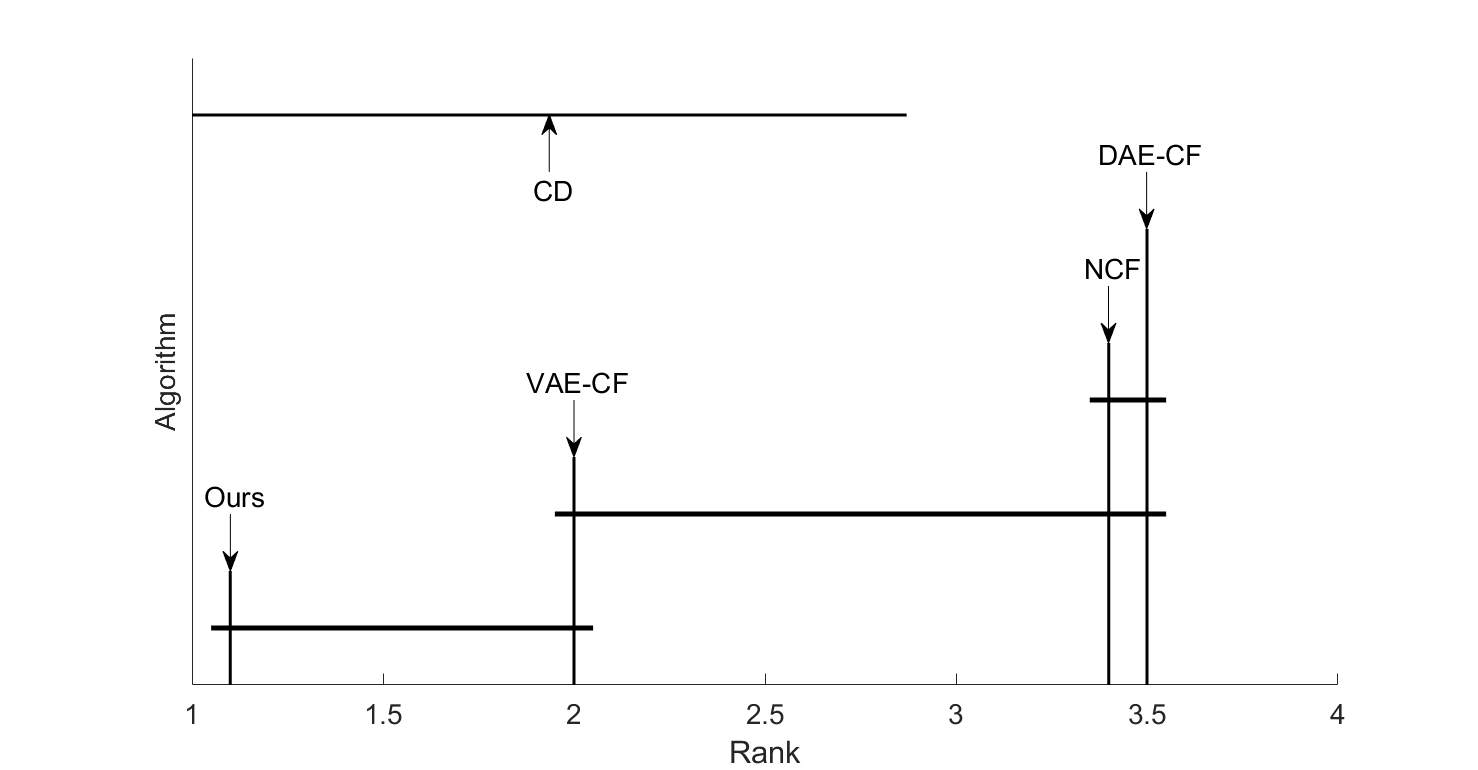}
        \caption{The CD diagram for Recall@50}~\label{fig:recall50_cd}
    \end{subfigure}
            \begin{subfigure}[t]{0.45\textwidth}
        \centering
        \includegraphics[height=1.3in]{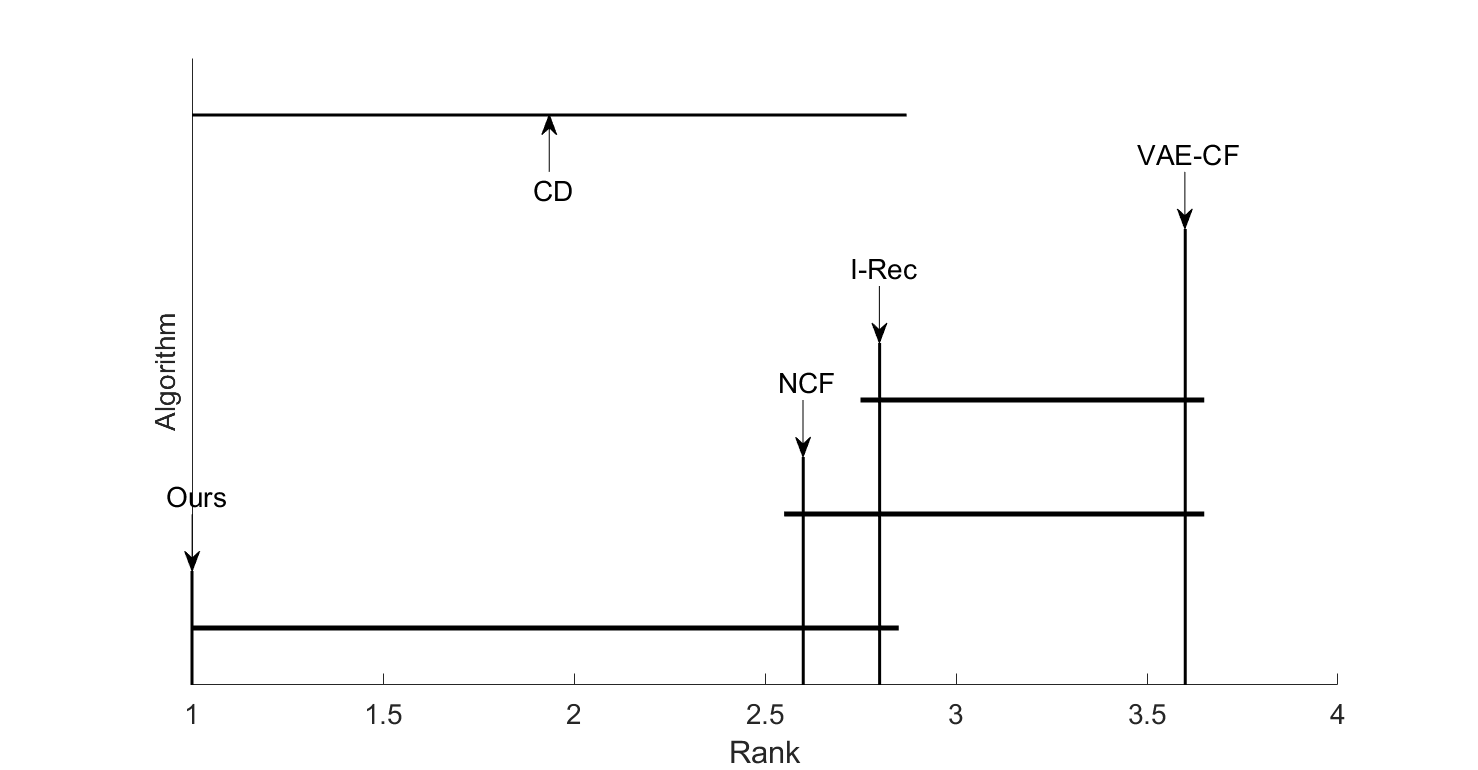}
        \caption{The CD diagram for RMSE}~\label{fig:rmse_cd}
    \end{subfigure}
    \caption{The critical difference (CD) diagrams showing the average ranks of the top 4 models on different evaluation metrics across the five datasets. The lower the rank (further to the left) the better performance of a model on the particular metric compared to the others on average. A line in each diagram indicates that there is no significant difference in performance among the models crossed by that particular line in terms of the Friedman test that compares the ranks of multiple classifiers~\citep{demvsar2006statistical}.}
\end{figure*}

To further compare the multiple models over the five datasets, we use critical difference (CD) diagram which is one of the standard evaluation tools in machine learning research~\citep{demvsar2006statistical,benavoli2016should}. We apply the Friedman test to compare the ranks of multiple models~\citep{demvsar2006statistical}. In this case, the null hypothesis states that there is no significant difference
in the mean rankings of the multiple models (at a statistical significant level 0.1\footnote{The level is set to be 0.1 to account for a small number of datasets in our experiment.}). Figure~\ref{fig:rmse_cd} shows the mean ranks of different models on RMSE; ours leading at 1, while NCF and I-Rec rivals at 2.58 and 2.79 respectively (statistically no different). It can also be seen that our framework has statistically different RMSE results from VAE-CF\footnote{We compare only the top 4 models to keep the number of models smaller than the number of datasets.}.

In terms of the Recall and the NDCG metrics, our framework performs better than or equal to any other baseline models across all the datasets. More specifically, on the ML-1M, ML-10M and Amazon-Kindle datasets, our framework consistently outperforms all the baseline models in all these metrics. On the ML-20M dataset, our framework ties with the VAE-CF model in the Recall@50 while on the Amazon-Pet dataset, it ties with the same model in the Recall@20. Another interesting finding is that the I-Rec model, despite its superiority in the RMSE on the Movielens datasets, is significantly inferior to the other models across all the datasets under these user-oriented metrics. This is not unusual as the I-Rec model focuses on modelling the characteristics of each item rather than those of the individual users. In comparison, the performance of models like the U-Rec, VAE-CF and DAE-CF, which are designed to capture users' underlying characteristics, is much less affected when evaluated under the user-oriented metrics. It can also be observed that models which have much more advantage in performance under the RMSE metric (e.g. our framework and the NCF) generally have less to no advantage compared to user-oriented models (e.g. VAE-CF and DAE-CF) under the NDCG and Recall metrics (see Figures~\ref{fig:ndcg20_cd},~\ref{fig:recall20_cd} and~\ref{fig:recall50_cd}). This is largely due to the fact that these models are trained to minimize the MSE loss rather than optimize the ranking of relevant items. With that being said, our framework manages to yield significantly different NDCG@50 performance from the other top models (see Figure~\ref{fig:ndcg50_cd}). 

\begin{figure*}[t]
%\begin{subfigure}[b]{0.7\columnwidth}
\centering
\includegraphics[width=5in]{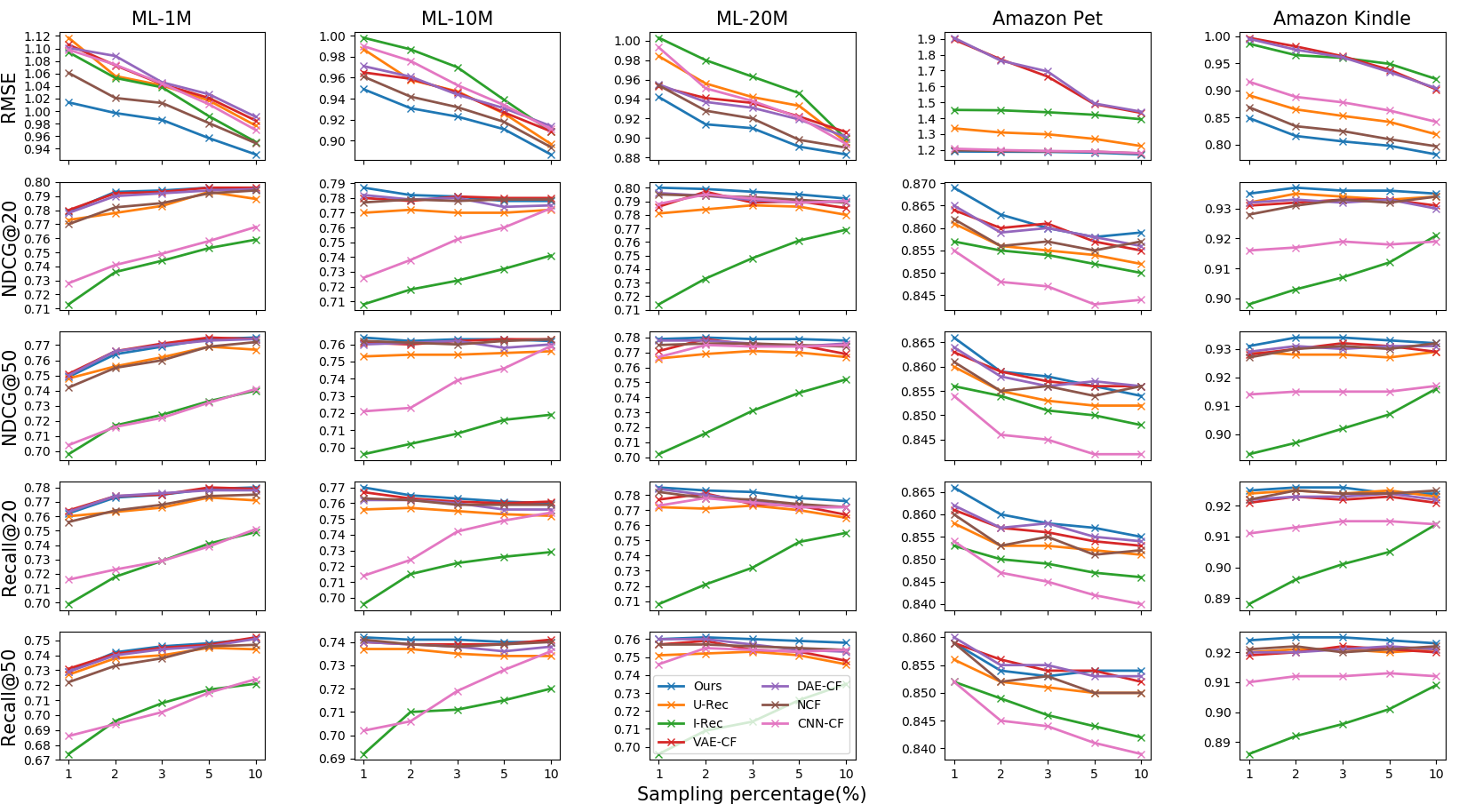}
  %\caption{Ability-Difficulty-Response Interaction}  
  \caption{The evaluation results of each model on the sub-sampled datasets under different metrics; The X-axis shows the sampling percentages which are 1\%, 2\%, 3\%, 5\% and 10\%. The Y-axis shows the testing RMSE.}
  \label{fig:subsampled_rmse}
\end{figure*}

As a further evaluation, we would like to see how robust our framework is towards the data sparsity issue, i.e., small numbers of responses per user and item, which often happens in the cold-start scenario~\citep{rashid2008learning}. We sub-sampled all the datasets randomly (irrespective of users and items) to make them sparse across both users and items. The sub-sampling percentages we have taken are $1\%, 2\%, 3\%, 5\%$ and $10\%$. The sub-sampled datasets are used to train each model. The rest of the data is split equally for optimizing the model hyper-parameters (with the validation datasets) and testing. Figure~\ref{fig:subsampled_rmse} shows the changes of the testing RMSE, NDCG and Recall scores of each model across different sub-sampling percentages. It can be observed that overall, our framework either outperforms or performs equally well against all the baselines in terms of these metrics on each dataset. The results in this case are largely consistent with the previous results on the full datasets; Our framework clearly performs better than the baselines in terms of the RMSE. The smallest margin in this case is 0.005 on the Amazon-Pet dataset when the sub-sampling percentage is 3\%. Its advantage in performance becomes relatively less under the ranking-type metrics most likely because of it minimizing the MSE loss. As for the baselines, the NCF model ranks at the second place in terms of the testing RMSE, while it is the VAE-CF model that ranks at this position in terms of the testing NDCG and Recall. It also appears that the I-Rec model suffers the most from the sparsity issue when evaluated by the ranking-type metrics.

%The best explanation to the robustness of the NCF towards the user and item rating sparsity on RMSE is that the MLP network takes in dense embeddings as the input rather than take in the sparse ratings directly (as by the I-AutoRec and U-AutoRec). The VAE-CF, DAE-CF and U-AutoRec models are generally under-performing for both the full and the sub-sampled datasets in terms of RMSE. In comparison, they perform much better on the NDCG and Recall which coincide with their intrinsic user-oriented nature.

\subsection{Ablation Study}
In this section, we conduct an ablation study to illustrate the importance of the three main components of our framework: the observed data inputs, the cross-feedback and the attention mechanisms. The observed data inputs include the rating matrix $\boldsymbol{X}$ and its transpose $\boldsymbol{X}^{T}$ that are fed to the user-side and item-side encoders respectively. Removing them and the entire MLP encoders that generate the corresponding intermediate embeddings from the framework means that the explicit data information is no longer exploited. The cross-feedback and attention components are based on the two latent inputs $\boldsymbol{Z}_{\boldsymbol{U}}$ and $\boldsymbol{Z}_{\boldsymbol{V}}$. Removing these components from our framework means that it no longer leverages the implicit data information for the variational inference.

With each of its components taken away, we train and optimize the hyper-parameters of our framework with the same training and validation datasets. Afterwards, we evaluate it on the same testing datasets in terms of all the metrics. Table~\ref{tab:results_rmse_ablation} shows the study results. It can be observed that all the components are important to our framework with respect to all the metrics. This is because when each component is removed, the corresponding performance of our framework degrades notably. Especially, the degradation caused by removing the data inputs is much more greater than that caused by removing the cross-feedback and attention mechanisms. This is expected as the explicit data information is more useful in improving the model performance compared to the implicit information. In addition, removing the attention mechanism in general causes the least degradation in performance; which still shows its importance as it further boosts the performance of our framework to surpass different baselines on different datasets. Overall, the ablation study shows the efficacy of the cross-feedback and attention mechanisms in improving the performance of our framework.

\begin{table}[t]
\centering
\scriptsize
  \begin{tabular}{llcccccc}
    \toprule
     & Components& ML-1M & ML-10M & ML-20M & ML-Pet & ML-Kindle\\
    \midrule
     &\textbf{No data input}& 0.917  & 0.902  &   0.878 & 1.241 &0.813\\
     &\textbf{No cross feedback}  & 0.864 & 0.829 & 0.824 & 1.110 &0.739\\
     \text{RMSE}&\textbf{No attention}  & 0.852 & 0.812 & 0.809 & 1.087 &0.718\\
     %\hline
          &\textbf{Full Model}& \textbf{0.842} & \textbf{0.804} & \textbf{0.798}  & \textbf{1.074} & \textbf{0.702}\\
\hline
    &\textbf{No data input} & 0.795 & 0.797 &0.804  & 0.907  & 0.961\\
    &\textbf{No cross feedback} & 0.798 & 0.801 &0.807  & 0.908 & 0.963\\
    NDCG@20&\textbf{No attention} & 0.800  & 0.803 &0.808   & 0.910 & 0.965\\
    &\textbf{Full Model} & \textbf{0.802} & \textbf{0.806} & \textbf{0.809}  & \textbf{0.911} & \textbf{0.966}\\
        \hline
    &\textbf{No data input} &  0.787 & 0.790 & 0.793  & 0.906 & 0.961\\
    &\textbf{No cross feedback} & 0.790 & 0.793  & 0.796 & 0.907 & 0.963\\
    NDCG@50&\textbf{No attention} & 0.792 & 0.795 & 0.798  & 0.909  & 0.964\\
    &\textbf{Full Model} & \textbf{0.793} & \textbf{0.797} &\textbf{0.800}  & \textbf{0.910} &  \textbf{0.965}\\      
    \hline
    &\textbf{No data input} & 0.775 & 0.776 & 0.786 & 0.902  & 0.956\\
    &\textbf{No cross feedback} & 0.778 & 0.779 &0.788  &0.904  & 0.958\\
    Recall@20&\textbf{No attention} & 0.780 & 0.782 &0.789  & 0.905 & 0.960\\
    &\textbf{Full Model} & \textbf{0.781} & \textbf{0.784} & \textbf{0.790} & \textbf{0.906} & \textbf{0.961}\\  
    \hline
    &\textbf{No data input} & 0.764 & 0.767 & 0.781 & 0.901 & 0.955\\
    &\textbf{No cross feedback} & 0.767 & 0.771 &0.782  & 0.903 & 0.957\\
    Recall@50&\textbf{No attention} & 0.768 & 0.773 & 0.782 & 0.905 & 0.959\\
    &\textbf{Full Model} & \textbf{0.770} & \textbf{0.774} & \textbf{0.783}  & \textbf{0.906} & \textbf{0.961}\\ 
    \bottomrule
  \end{tabular}
    \caption{The ablation study shows that all components contribute significantly to the framework performance in terms of the RMSE.}
  \label{tab:results_rmse_ablation}
\end{table}

\subsection{Convergence analysis}

To further investigate the importance of the cross-feedback and attention mechanism, we inspect how the framework performance (i.e. the testing RMSE scores) converges with and without the mechanism. More specifically, we inspect the performance of the following five variants of our framework: 
\begin{itemize}
    \item \textbf{Variant A}: our framework \textbf{without} the cross-feedback mechanism, thereby no latent input, and the number of hidden layers is $L'=0$ for the MLP networks over the observed input.
    \item \textbf{Variant B}: our framework \textbf{with} the \textbf{cross-feedback} (but no attention) mechanism and the number of hidden layers (before ``concat'') is $L'=0$ for the MLP networks over both the observed and the \textbf{latent} inputs. The number of hidden layers (after ``concat'') is $L=0$.
    \item \textbf{Variant C}: our framework \textbf{without} the cross-feedback mechanism and the number of hidden layers is $L'=1$ for the MLP networks over the observed input.
    \item \textbf{Variant D}: our framework \textbf{with} the \textbf{cross-feedback} (but no attention) mechanism and the number of hidden layers (before ``concat'') is $L'=1$ for the MLP networks over both the observed and the \textbf{latent} inputs. The number of hidden layers (after ``concat'') is $L=0$.
    \item \textbf{Variant E}: our framework \textbf{with} the \textbf{cross-feedback} and the \textbf{attention} mechanisms. The number of hidden layers (before ``concat'') is $L'=1$ for the MLP networks over both the observed and the \textbf{latent} inputs. The number of hidden layers (after ``concat'') is $L=0$.
\end{itemize}

%Here, the only difference between variants A and B, and between variants C and D is the presence of the cross-feedback mechanism. We would like to see whether variants B and D which possess the mechanism will not only perform better but also converge faster than their counterparts without the mechanism.

\begin{figure}[h]
%\begin{subfigure}[b]{0.7\columnwidth}
\centering
\includegraphics[width=2in]{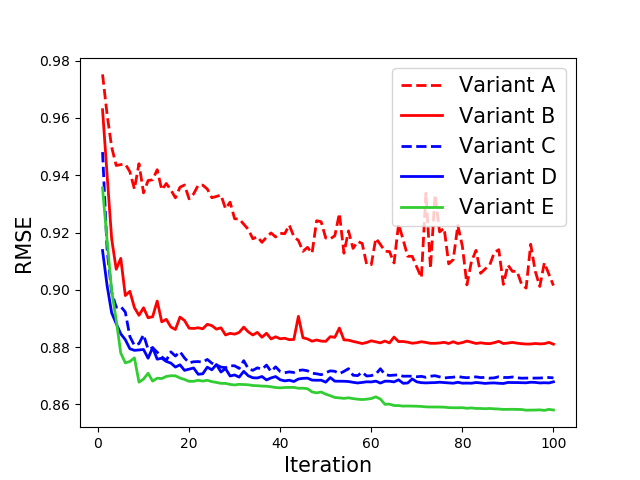}
  %\caption{Ability-Difficulty-Response Interaction}  
  \caption{The testing RMSE scores of the five variants across the first 100 training iterations over the ML-1M data. The X-axis shows the iteration number, while the Y axis shows the testing RMSE scores.}

\label{fig:convergence_analysis}
\end{figure}

Figure~\ref{fig:convergence_analysis} shows how the testing RMSE scores of the five variants change during the first 100 training iterations over the ML-1M dataset. It can be observed that the three variants (i.e. variants B and D) with the cross-feedback mechanism constantly outperform their counterparts without the mechanism (i.e. variants A and C). Especially, the RMSE score of variant D declines much more quickly than that of variant C at the early stage of the training. This indicates that the cross-feedback mechanism can quickly improve the framework performance even when the posterior inference for the embeddings has just begun. We also notice that the line of variant B is much smoother than that of variant A which still fluctuates after 50 iterations. Considering the two variants have the same optimization configuration, this clearly shows that the cross-feedback mechanism smooths and accelerates the convergence of our framework. Furthermore, there is a huge performance gap between variants A and B. This gap reconfirms that the cross-feedback mechanism can significantly improve the framework performance when the explicit data information is not sufficiently exploited (due to no hidden layer in the corresponding MLPs). Finally, when the attention mechanism is integrated into our framework (see variant E in the figure), the framework's performance increases significantly and converges almost steadily after 20 iterations. It is also worth noting that the attention mechanism slows down the performance improvement (created by the cross-feedback mechanism) at the early stage of the training. This is most likely because the learning of the additional model parameters $\boldsymbol{\Theta}$ and $\boldsymbol{\Lambda}$ (at each side) is tangled with the learning of the embeddings, which is more difficult than the learning of the embeddings alone.

As for the running time of our framework, which includes the model induction and prediction time, it is largely determined by the input dimension which is $\mathcal{O}\big((K+1)\times(I+J)\big)$ for our framework. Inevitably, this results in our framework being notably slower than simpler models such as VAE-CF, DAE-CF, U-Rec and I-Rec in exchange for improved performance. With that being said, thanks to the generally fast convergence, our framework is in general just 4-5 times slower than these models on large-scale datasets such as ML-20M and Amazon datasets; which have large numbers of users and items. We found empirically that the running time of our model overall rivals that of the NCF and CNN-CF models. We have also experimented with an accelerated variant of our framework which, instead of taking in a concatenation of the embeddings, uses their \textbf{average} as the latent input. However, we observed that this variant constantly yielded worse performance, indicating that it might fail to fully exploit the implicit information. 

\section{Conclusion}
In this paper, we propose a variational auto-encoder based Bayesian matrix factorization framework for collaborative filtering. Compared to the previous work, this framework leverages not only the explicit data information from the ratings but also the implicit information from the user and item embeddings for improving the variational approximation of their posterior parameters.

Our framework is characterized by the iterative cross feedback of the user and item embeddings to each other's encoder input layers. The cross-feedback inputs provide useful implicit information regarding the counterparts. Both the explicit and implicit information is learned by dedicated MLP networks whose final outputs are concatenated. Then, the result is fed into another MLP network which fuses and map the two types of information into the posterior parameters of the embeddings. To better exploit the implicit information provided by the cross-feedback component, we propose to further leverage the attention of the posterior parameters of one side to the embeddings cross fed from the other side.

Experimental results show that our framework outperforms six state-of-the-art NN based collaborative filtering approaches in terms of the reconstruction error over both the full and sub-sampled data. They also show that our framework performs no worse than those baselines with respect to the ranking of relevant items in both scenarios. It is further found from the ablation study and the convergence analysis that both the cross-feedback and the attention mechanisms not only improve the framework performance but also accelerate its convergence even though the attention mechanism slows down such improvement at the beginning of the training. As the future work, we would like to investigate how to accelerate the running time of our framework by improving its latent input structure.

%As the future work, we will investigate the possibility of incorporating the attention mechanisms and the transformers into the encoder structure to refine the posterior parameter estimation for the embeddings. We will also adapt our framework to be able to model the implicit data in recommendation systems.

%\begin{table}[t]
%\centering
%  \begin{tabular}{lcccccc}
%    \toprule
%     & 1M & 10M & 20M & Pet & Kindle\\
%    \midrule
%    \textbf{NCF} & 0.870  & 0.831  & 0.822  & \underline{1.098}  &\underline{0.734}\\
%        \textbf{CNN-CF} & 0.875 & 0.844  & 0.839  & 1.119  &0.770\\
%    \textbf{U-Rec} & 0.883  & 0.893 & 0.833  & 1.136  &0.772\\
%    \textbf{I-Rec} & \underline{0.859}  & \underline{0.823} & \underline{0.818}  & 1.213  &0.787\\
%    \textbf{DAE-CF} & 0.911 &  0.852 & 0.841  & 1.247  & 0.762\\
%    \textbf{VAE-CF} & 0.903 & 0.834  & 0.836  &1.211  &0.747\\
%    \textbf{Rec-VAE} &0.986&&&&\\
%    \textbf{Ours} & \textbf{0.848} & \textbf{0.812} & \textbf{0.809}  & \textbf{1.087} & \textbf{0.718}\\
%    \bottomrule
%  \end{tabular}
%  \caption{The RMSE results of each model across the five datasets.}
%  \label{tab:results_rmse_full}
%\end{table}

\bibliographystyle{spbasic}
\bibliography{bibliography} 

\end{document}